%% file: main.tex
\documentclass[11pt]{article}

\usepackage[preprint]{acl}

\usepackage{times}
\usepackage{latexsym}

\usepackage[T1]{fontenc}
\usepackage[utf8]{inputenc}

\usepackage{microtype}

\usepackage{inconsolata}

\usepackage{graphicx}

\usepackage{hyperref}       % hyperlinks
\usepackage{url}            % simple URL typesetting
\usepackage{booktabs}       % professional-quality tables
\usepackage{amsfonts}       % blackboard math symbols
\usepackage{nicefrac}       % compact symbols for 1/2, etc.
\usepackage{microtype}      % microtypography
\usepackage{xcolor}         % colors

\usepackage[most]{tcolorbox}
\usepackage{makecell}
\usepackage{hyperref}
\usepackage{url}
\usepackage{bbm}
\usepackage{caption}
\usepackage{quickmath}
\usepackage{lipsum}
\usepackage[ruled,vlined]{algorithm2e}
\SetKwComment{Comment}{/*}{*/}
\usepackage{array}
\usepackage{soul}
\usepackage{titletoc}
\usepackage{subcaption} 
\usepackage{fontawesome5}

\usepackage[table]{xcolor}
\definecolor{lightpurple}{RGB}{180,200,230}
\definecolor{linkblue}{RGB}{70,130,180}
\sethlcolor{lightpurple!25}
\usepackage{enumitem,amssymb}
\usepackage{wrapfig}
\usepackage{multirow}  % For multirow support
\usepackage{graphicx}
\newlist{todolist}{itemize}{2}
\setlist[todolist]{label=$\square$}
\usepackage{pifont}
\usepackage{multirow}
\usepackage{tabularx,threeparttable,array}
\usepackage{algorithmic}
\usepackage{xspace}
\usepackage{subcaption}

\newcolumntype{P}{>{\centering\arraybackslash}X}

\newcolumntype{Y}{>{\centering\arraybackslash}X}

\newcolumntype{Z}{>{\columncolor{lightpurple!25}\centering\arraybackslash}X}

\newcommand{\diff}[1]{\textcolor{black}{#1}}

\newcommand{\ours}{\textsc{GeoSteer}\xspace}

\renewcommand{\arraystretch}{1}

\title{\textbf{\ours{}}: Geodesic Optimization for Activation Steering in Large Language Models}

\author{Xuan Cuong Ngo \qquad Hao Vo \qquad Ngan Le \\
        University of Arkansas, Fayetteville, Arkansas, USA \\
        {\tt\small \{cngo,haov,thile\}@uark.edu}}

\begin{document}
\maketitle

\input{content/abstract}
\input{content/intro}

\input{content/related_work}
\input{content/method_0}

\input{content/exp}
\input{content/concl}

\bibliography{custom}

\appendix

\input{content/appendix}
% \section{Example Appendix}
% \label{sec:appendix}

% This is an appendix.

\end{document}

%% file: content/abstract.tex
\begin{abstract}
% Activation steering is a lightweight method to control the behaviour of Large Language Model (LLM) by controlling the hidden activation of the model. Norm-preserving is a family of method to steer without trying to change to norm of activation to avoid representation collapse and degradation. However, current methods suffer from two key limitations: (i) their steering trajectories are largely predefined (ii) an over-reliance on one-step steering that fail to capture complex patterns of activation distributions. In this work, we propose a optimization-based method to steer activation for LLM. We view the activation steering as a riemanian optimization problem. Based on this perspective, we view steering as stepping many small steps toward geodesic update in the curved space and identifying the direction is by following the trajectories that optimize a barrier function. GeoSteer identifies steering directions by defining the barrier function as the log-density ratio between positive and negative activations, and employs it to construct an ODE for multistep and adaptive steering in the Riemanian manifold. This recursive formulation yields more stable and consistent steering dynamics. Compared to state-of-the-art activation steering methods, GeoSteer achieves consistent empirical improvements on diverse LLM alignment benchmarks of TruthfulQA,  RealToxicityPrompts and UltraFeedback. This work establishes a principled new view of norm-preserving activation steering in LL by unifying optimzation and steering method.
Activation steering provides a lightweight way to control large language models (LLMs) by modifying their hidden activations at inference time. Among these approaches, norm-preserving steering aims to change model behavior without altering the activation norm, reducing the risk of representation collapse and degradation. However, existing norm-preserving methods are limited by predefined steering trajectories and by their reliance on one-step updates, which may fail to capture the complex structure of activation distributions.
We propose \textbf{\ours{}}, an optimization-based method for \emph{norm-preserving activation steering}. \ours{} formulates steering as a Riemannian optimization problem and updates activations through a sequence of small geodesic steps on the representation manifold. To avoid fixed steering directions, 
% \ours{} defines an objective function based on the log-density ratio between positive and negative activations,
\ours{} learns a nonlinear activation-space objective that distinguishes desired from undesired activations,
and uses this function to adaptively guide each steering step. This multistep formulation yields smoother, more stable, and more consistent steering behavior while preserving the activation norm.
Across TruthfulQA, RealToxicityPrompts, and UltraFeedback benchmarks, \ours{} consistently improves over state-of-the-art activation steering baselines. These results suggest that norm-preserving steering can be made more effective by replacing predefined one-step edits with adaptive, geometry-aware optimization.
\end{abstract}

%% file: content/intro.tex
\section{Introduction}

\begin{figure*}[t]
    \centering
    \includegraphics[width=\linewidth]{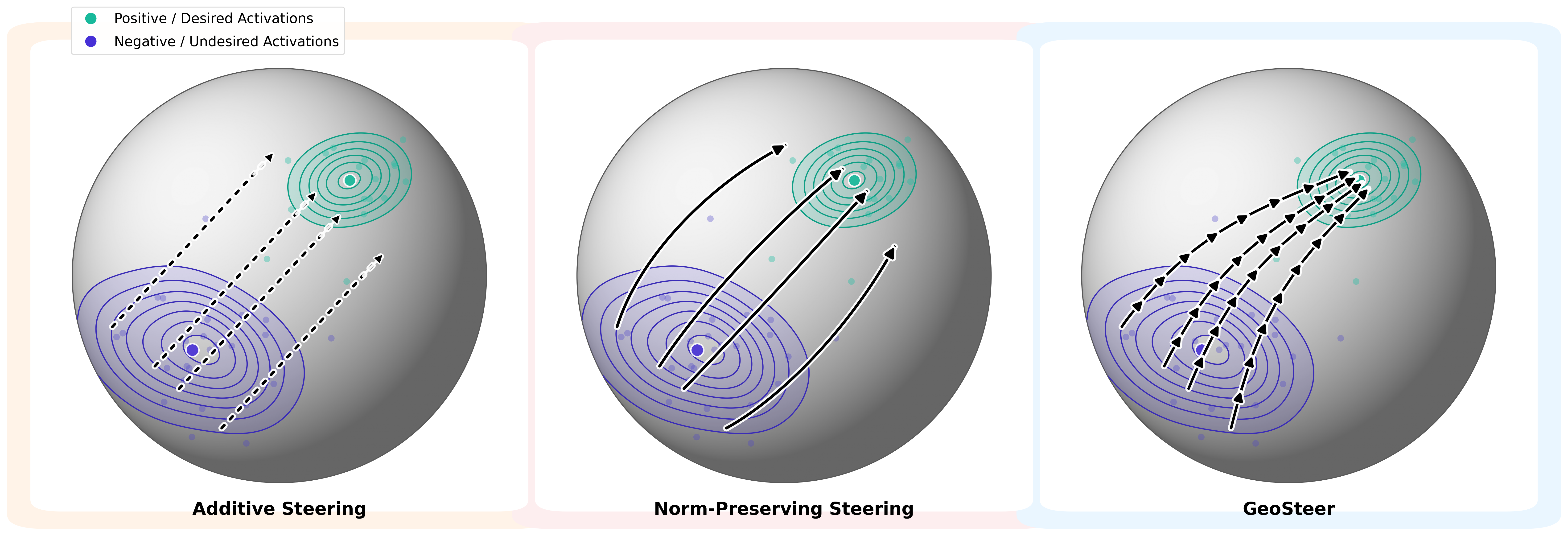}
    % \caption{Blue regions denote negative/undesired activations, while green regions denote positive/desired activations. Additive steering applies a fixed Euclidean shift to hidden activations; the dotted arrows indicate that this update is not constrained to the sphere surface and may move through the interior of the activation manifold. Norm-preserving steering keeps activations on the sphere but uses a single geodesic-like update, which can miss the desired activation region when the same global direction is insufficient. In contrast, GeoSteer formulates steering as Riemannian optimization on the norm-preserving activation manifold, decomposing the intervention into multiple small geodesic updates guided by the current activation state. This produces adaptive trajectories that gradually move undesired activations toward the desired distribution while preserving activation norm.}
    \caption{
    Conceptual illustration of \ours{}. Additive steering can move activations off the sphere, while existing norm-preserving steering keeps activations on the sphere but typically follows predefined or heuristic trajectories. In contrast, \ours{} adaptively optimizes the steering path by decomposing the intervention into multiple geodesic updates. This produces smoother trajectories toward the desired activation region while preserving the activation norm.
    }
    \label{fig:teaser}
\end{figure*}

Large language models (LLMs) have demonstrated remarkable capabilities and have become an essential component of modern AI, supporting a wide range of applications \citep{achiam2023gpt,dubey2024llama,liu2024deepseek,yang2025qwen3}. However, as LLMs continue to develop rapidly, it is increasingly important to study and control their societal impacts, including hallucination, bias, toxicity, and other harmful behaviors. Since modern LLMs often contain hundreds of billions of parameters \citep{chowdhery2023palm,brown2020language}, controlling and aligning them through fine-tuning is becoming increasingly challenging and resource-intensive. \emph{This motivates the need for lightweight alternatives that can adapt model behavior without updating model parameters.}

% Activation steering, also known as representation engineering, has emerged as a simple and effective approach for controlling LLM behavior ~\citep{rimsky2024steering, wehner2025taxonomy, bartoszcze2025representation}. Most existing methods perform steering through activation addition, where a vector derived from contrastive examples is linearly injected into hidden activations during decoding ~\citep{li2023inference,rimsky2024steering,liu2023context,lee2024programming}. Despite its simplicity, activation addition is highly sensitive to scale: small steering offsets often have limited effect, while large offsets can disrupt the generation distribution and degrade model quality \cite{you2026spherical}. Recent learnable intervention methods suggest that stronger and more reliable control requires more structured and geometry-aware mechanisms rather than unconstrained linear shifts \citep{wu2024reft,deng2026steering}. Meanwhile, modern LLM architectures provide a natural geometric prior: normalization layers such as RMSNorm \citep{zhang2019root} stabilize activation magnitudes, making activation direction a more meaningful degree of freedom for steering.

Activation steering, or representation engineering, offers a lightweight way to influence LLM behavior by modifying hidden states at inference time~\citep{rimsky2024steering, wehner2025taxonomy, bartoszcze2025representation}. A common strategy is to construct a steering vector from contrastive examples and add it to model activations during generation~\citep{li2023inference,rimsky2024steering,liu2023context,lee2024programming}. While effective in many settings, such additive edits are highly sensitive to the intervention scale. If the intervention is too weak, the model behavior changes little; if it is too strong, the edited activation may move away from the model's natural representation space and harm generation quality~\citep{you2026spherical}. This limitation has motivated recent work on more structured intervention mechanisms, including learnable and geometry-aware approaches~\citep{wu2024reft,deng2026steering}. At the same time, the architecture of modern LLMs suggests a useful geometric perspective: normalization layers such as RMSNorm~\citep{zhang2019root} regulate activation scale, making the direction of the hidden representation an important target for controlled steering.

% Norm-preserving steering methods have therefore been proposed as a way to control model behavior while maintaining the activation norm \citep{pham2024hpr,you2026spherical,vu2026angular}. However, existing norm-preserving methods still have important limitations. Many methods rotate activations along predefined or heuristic trajectories rather than adaptively optimizing the steering path. For example, one-step approaches such as Angular Steering \cite{vu2026angular} and Spherical Steering \cite{you2026spherical} often rely on simple statistical features, such as mean activation differences, while ignoring richer distributional structure and interactions among activation dimensions. Their reliance on one-step updates may also fail to capture the complex patterns of activation distributions, limiting their expressive power when steering nuanced model behaviors.

Norm-preserving steering methods have therefore been proposed as a way to control model behavior while maintaining the activation norm~\citep{pham2024hpr,you2026spherical,vu2026angular}. However, existing norm-preserving methods still have important limitations, as conceptually illustrated in Figure~\ref{fig:teaser}. While they avoid the unconstrained Euclidean shifts used by additive steering, they often follow predefined or heuristic trajectories rather than adaptively optimizing the steering path. For example, one-step approaches such as Angular Steering~\citep{vu2026angular} and Spherical Steering~\citep{you2026spherical} typically rely on simple statistical features, such as mean activation differences, while ignoring richer distributional structure and interactions among activation dimensions. Their reliance on one-step updates may also fail to capture the complex patterns of activation distributions, limiting their expressive power when steering nuanced model behaviors.

Other methods, such as HPR \cite{pham2024hpr}, introduce more complex training procedures involving neural networks, but still assume that the correct steering direction lies within a predefined two-dimensional plane. This geometric restriction can limit flexibility, especially when the desired behavioral change requires movement along more complex directions in representation space. Moreover, existing methods are often sensitive to hyperparameters and may not generalize consistently across different models or tasks.

To address these limitations, we propose \ours, an optimization-based framework for norm-preserving activation steering. We observe that norm-preserving steering can be naturally viewed as moving an activation point on a Riemannian manifold from an undesired region toward a desired region, while keeping its norm fixed. Rather than applying a single large update or following a predefined rotation, \ours decomposes steering into a sequence of small geodesic updates \cite{absil2008optimization}. Each update is guided by the current activation state, and together these steps form a smooth, adaptive trajectory on the manifold.
We formulate this process as a Riemannian optimization problem \cite{absil2008optimization}. 
% Specifically, we define a steering objective using the log-density ratio between desired and undesired activation distributions, modeled through nonlinear features.
Specifically, we define a nonlinear activation-space objective that distinguishes desired from undesired activations.
This objective provides an adaptive signal for selecting steering directions that increase alignment with desired activations while moving away from undesired ones. At inference time, \ours performs iterative geodesic updates guided by this objective, ensuring that activations remain on the norm-preserving manifold throughout the steering process.
By replacing predefined one-step rotations with adaptive multistep optimization, \ours captures richer and more fine-grained structure in the activation space. This leads to more stable and expressive steering while preserving activation norms. Experiments on TruthfulQA \cite{lin2021truthfulqa}, UltraFeedback \cite{cui2023ultrafeedback}, and RealToxicityPrompts \cite{gehman2020realtoxicityprompts} show that \ours consistently outperforms state-of-the-art activation steering baselines.

Our contributions are summarized as follows:
\begin{itemize}[topsep=0pt, partopsep=0pt, parsep=0pt]
    \item We formulate norm-preserving activation steering as Riemannian optimization on the activation sphere, providing a principled geometric view of activation intervention.

    \item We propose \ours{}, an adaptive multistep steering method that replaces fixed one-step transformations with a sequence of small geodesic updates guided by the current activation state.

    \item We conduct experiments on TruthfulQA, UltraFeedback, and RealToxicityPrompts, showing that \ours{} consistently improves over strong activation steering baselines while maintaining efficient inference.
\end{itemize}

%% file: content/related_work.tex
\section{Related Work}
% \textbf{Activation Steering.} Activation steering controls LLM behavior by modifying hidden activations at inference time. Most existing methods use additive steering, where a steering vector is injected into the activation space. Fixed-vector methods such as RepE \cite{zou2023representation}, ITI \cite{li2023inference}, and CAA \cite{rimsky2024steering} apply the same update across inputs, which limits their adaptability. Other extensions, such as MiMiC \cite{singh2024representation} and Linear-AcT \cite{rodriguez2025controlling}, introduce more structured formulations but still rely on restrictive linear assumptions. However, additive methods can push activations away from their natural representation distribution, leading to model collapse and degraded utility.
\textbf{Activation Steering.}
Activation steering controls LLM behavior by modifying hidden activations at inference time. Most existing methods follow an additive steering paradigm, where a direction vector is estimated from contrastive activation patterns and added to the model's hidden states during decoding. The intervention typically takes the form of shifting an activation by a scaled steering vector, so that the edited representation is moved toward a desired behavioral direction in activation space. Fixed-vector methods such as ITI~\cite{li2023inference}, RepE~\cite{zou2023representation}, 
and CAA~\cite{rimsky2024steering} apply the same update across inputs, which limits their adaptability. MiMiC~\cite{singh2024representation} and Linear-AcT~\cite{rodriguez2025controlling} introduce more structured linear formulations, while ODESteer~\cite{zhao2026odesteer} further improves adaptivity by treating steering as an ODE-based trajectory. \emph{Nevertheless, these approaches do not explicitly constrain the activation norm, so their updates may still push representations away from the model's natural activation geometry and degrade utility.}

\noindent
\textbf{Norm-Preserving Activation Steering.} Norm-preserving activation steering addresses this issue by modifying activation directions while keeping activation norms unchanged. These methods are motivated by the observation that modern LLM activations often lie on structured manifolds, where changing the direction can influence model behavior without disrupting activation magnitude. Angular Steering \cite{vu2026angular} and HPR \cite{pham2024hpr} perform geometric updates using fixed low-dimensional planes or Householder reflection, but they still impose strong assumptions on the steering direction. Spherical Steering \cite{you2026spherical} rotates activations on the sphere, but often relies on simple mean-based statistics and ignores richer distributional structure. Moreover, most norm-preserving methods perform only a one-step update, limiting their ability to capture complex activation patterns. \emph{In contrast, our method performs adaptive multistep updates on the manifold, recomputing the steering direction at each step based on the current activation.}

\textbf{}

%% file: content/method_0.tex
\section{Methodology}

\subsection{Preliminaries}

\textbf{Activation Steering.}
Activation steering aims to modify an internal hidden representation of a pretrained language model so that the model behavior shifts toward a desired attribute while preserving the original model parameters. Let $h \in \mathbb{R}^d$ denote the activation at a selected layer and token position. Conventional steering methods often edit the activation by adding a direction vector,
\begin{equation}
\hat{h} = h + \alpha v,
\end{equation}
where $v$ is a steering direction and $\alpha$ controls the steering strength. While simple, this additive update may change both the direction and norm of the activation, which can move the representation away from the geometry learned by the model.

% \noindent
% \textbf{Norm-Preserving Steering on the Hypersphere.}
% Recent norm-preserving steering methods constrain the edited activation to remain on a hypersphere with the same norm as the original activation. Given an activation $h$, we first write its normalized representation as

% \begin{equation}
% z = \frac{h}{\|h\|_2}, \quad z \in \mathbb{S}^{d-1}.
% \end{equation}

% Steering is then performed on the unit sphere by modifying $z$ into an edited normalized activation $\hat{z} \in \mathbb{S}^{d-1}$. The final edited activation in the original activation space is recovered by restoring the original norm:

% \begin{equation}
% \hat{h} = \|h\|_2 \hat{z}.
% \end{equation}

% Here, $\hat{z}$ denotes the steered direction on the unit sphere, while $\hat{h}$ denotes the final edited activation used by the model. This formulation preserves the activation norm and treats steering as a directional update over the sphere rather than as an unconstrained Euclidean shift.

\noindent
\textbf{Norm-Preserving Steering on the Hypersphere.}
Norm-preserving steering constrains the edited activation to remain on a hypersphere with the same norm as the original activation. Given an activation $h$, its normalized representation can be written as
\begin{equation}
z = \frac{h}{\|h\|_2}, \quad z \in \mathbb{S}^{d-1}.
\end{equation}
Steering can then be viewed as modifying $z$ on the unit sphere to obtain an edited normalized activation $\hat{z} \in \mathbb{S}^{d-1}$. The final edited activation in the original activation space is recovered by restoring the original norm:
\begin{equation}
\hat{h} = \|h\|_2 \hat{z}.
\end{equation}
Here, $\hat{z}$ denotes the steered direction on the unit sphere, while $\hat{h}$ denotes the final edited activation used by the model. This formulation preserves the activation norm and treats steering as a directional update over the sphere rather than as an unconstrained Euclidean shift.

% \noindent
% \textbf{Riemannian Optimization on the Sphere.}
% The unit sphere $\mathbb{S}^{d-1}$ is a Riemannian manifold. Therefore, an update that stays on the sphere should move along the tangent space and then map the result back to the manifold. Let $\mathcal{L}(z)$ denote a general objective function defined over a normalized activation $z \in \mathbb{S}^{d-1}$. For a point $z \in \mathbb{S}^{d-1}$, the tangent space is

% \begin{equation}
% T_z \mathbb{S}^{d-1}
% =
% \{u \in \mathbb{R}^d : u^\top z = 0\}.
% \end{equation}

% Given the Euclidean gradient $g = \nabla_z \mathcal{L}(z)$, its Riemannian gradient on the sphere is obtained by projecting it onto the tangent space:

% \begin{equation}
% g_{\mathcal{R}}
% =
% g - (g^\top z)z.
% \end{equation}

% This tangent vector gives the local direction that changes the objective while respecting the spherical constraint. In GeoSteer, this general objective is later instantiated as the probe-based steering loss.

\noindent
\textbf{Riemannian Optimization on the Sphere.}
Steering on the hypersphere can be viewed as a constrained optimization problem. Given a normalized activation $z \in \mathbb{S}^{d-1}$ and an objective function $\mathcal{L}(z)$, the goal is to update $z$ so as to improve the objective while maintaining the constraint
\begin{equation}
\|z\|_2 = 1.
\end{equation}
A standard Euclidean gradient step does not generally satisfy this constraint, since it may move the activation away from the sphere. Riemannian optimization addresses this by computing the update direction in the tangent space of the manifold.

For the unit sphere, the tangent space at $z$ is
\begin{equation}
T_z \mathbb{S}^{d-1}
=
\{u \in \mathbb{R}^d : u^\top z = 0\}.
\end{equation}
This space contains all directions that are locally valid on the sphere. Given the Euclidean gradient
\begin{equation}
g = \nabla_z \mathcal{L}(z),
\end{equation}
we obtain the Riemannian gradient by removing the component of $g$ that points in the radial direction:
\begin{equation}
g_{\mathcal{R}}
=
g - (g^\top z)z.
\end{equation}
The resulting vector $g_{\mathcal{R}} \in T_z \mathbb{S}^{d-1}$ gives the steepest local change of the objective under the spherical constraint. After choosing a tangent update direction, the point can be moved along the sphere using a geodesic update. This provides the geometric basis for \ours{}, where $\mathcal{L}(z)$ is instantiated as a learned steering objective.

\subsection{\ours{}}
\label{sec:geosteer}

We consider inference-time activation steering for a pretrained language model. Given an input prompt $x$, let $h \in \mathbb{R}^d$ denote the activation at a selected layer and token position. \ours{} edits this activation by optimizing its normalized direction on the unit sphere. We denote the inference-time trajectory by $\{z^{(t)}\}_{t=0}^{K}$, initialized as
\begin{equation}
z^{(0)} = \frac{h}{\|h\|_2}.
\end{equation}
After $K$ geodesic steps, the edited activation is reconstructed as
\begin{equation}
\hat{h}=\|h\|_2 z^{(K)},
\end{equation}
so the original activation norm is preserved.

\noindent
\textbf{Learning the Steering Objective.}
\ours{} first learns a probe that distinguishes desired from undesired activations. Given labeled activations
\begin{equation}
\mathcal{D} = \{(h_i, y_i)\}_{i=1}^N,
\end{equation}
where $y_i \in \{0,1\}$ denotes the undesired or desired class, we normalize each training activation as 
\begin{equation}
z_i = \frac{h_i}{\|h_i\|_2}.
\end{equation}
The probe score is defined as: 
% $s_\phi(z) = w^\top \phi(z) + b,$
\begin{equation}
s_\phi(z) = w^\top \phi(z) + b,
\end{equation}
where $\phi(\cdot)$ is a nonlinear feature map, and $w$ and $b$ are learnable parameters. The desired-class probability is
\begin{equation}
p_\phi(z)=\sigma(s_\phi(z)).
\end{equation}
The probe is trained with the binary classification loss
\begin{equation}
\resizebox{0.95\columnwidth}{!}{$
\mathcal{L}_{probe}
=
-\sum_i
\left[
y_i \log p_\phi(z_i)
+
(1-y_i)\log(1-p_\phi(z_i))
\right].
$}
\end{equation}

In our implementation, $\phi(\cdot)$ is instantiated with Polynomial Count Sketch (PCS)~\cite{pham2013fast}:
\begin{equation}
\phi(z)=\mathrm{PCS}(z).
\end{equation}
This provides an efficient nonlinear representation without explicitly expanding high-dimensional polynomial features. After training, the probe is frozen and used to define the inference-time steering objective.

\noindent
\textbf{Geodesic Steering.}
At inference time, \ours{} steers the activation by maximizing the desired-class probability under the learned probe. Equivalently, we minimize
\begin{equation}
\mathcal{L}_{steer}(z) = -\log p_\phi(z).
\end{equation}
Starting from $z^{(0)}$, \ours{} performs $K$ geodesic steps with step size $\eta=T/K$, where $T$ is the total steering strength. At each step $t$, we compute the Euclidean gradient
\begin{equation}
g^{(t)} = \nabla_{z^{(t)}}\mathcal{L}_{steer}(z^{(t)}),
\end{equation}
and project it onto the tangent space of the sphere:
\begin{equation}
g_{\mathcal{R}}^{(t)}
=
g^{(t)} - \left((g^{(t)})^\top z^{(t)}\right)z^{(t)}.
\end{equation}
The normalized descent direction is
\begin{equation}
u^{(t)}
=
-\frac{g_{\mathcal{R}}^{(t)}}{\|g_{\mathcal{R}}^{(t)}\|_2}.
\end{equation}
\ours{} then updates the activation direction along the geodesic on the unit sphere \cite{absil2008optimization}:
\begin{equation}
z^{(t+1)}
=
\cos(\eta)z^{(t)}
+
\sin(\eta)u^{(t)}.
\end{equation}
Because this update follows the sphere, each intermediate point remains on $\mathbb{S}^{d-1}$. We provide the proof in Appendix~\ref{a:norm}. After $K$ steps, the edited activation is recovered as
\begin{equation}
\hat{h}=\|h\|_2 z^{(K)}.
\end{equation}

The full inference procedure is summarized in Algorithm~\ref{alg:geosteer}.

\begin{algorithm}[t]
\caption{\ours{} Inference}
\label{alg:geosteer}
\begin{algorithmic}[1]
\REQUIRE Activation $h$, frozen probe $p_\phi$, total steering strength $T$, number of steps $K$
\ENSURE Steered activation $\hat{h}$

\STATE $z^{(0)} \leftarrow h / \|h\|_2$
\STATE $\eta \leftarrow T / K$

\FOR{$t = 0$ to $K-1$}
    \STATE $\mathcal{L}_{steer}(z^{(t)}) \leftarrow -\log p_\phi(z^{(t)})$
    \STATE $g^{(t)} \leftarrow \nabla_{z^{(t)}}\mathcal{L}_{steer}(z^{(t)})$
    \STATE $g_{\mathcal{R}}^{(t)} \leftarrow g^{(t)} - \left((g^{(t)})^\top z^{(t)}\right)z^{(t)}$
    \STATE $u^{(t)} \leftarrow -g_{\mathcal{R}}^{(t)} / \|g_{\mathcal{R}}^{(t)}\|_2$
    \STATE $z^{(t+1)} \leftarrow \cos(\eta)z^{(t)} + \sin(\eta)u^{(t)}$
\ENDFOR

\STATE $\hat{h} \leftarrow \|h\|_2 z^{(K)}$
\RETURN $\hat{h}$
\end{algorithmic}
\end{algorithm}

\subsection{Comparison with Existing Norm-Preserving Steering.}
\label{sub:compare}

\ours{} differs from prior norm-preserving steering methods in two main aspects. First, rather than using a fixed geometric transformation, it treats activation steering as an optimization problem on the activation sphere. Second, the steering direction is determined by the local gradient of a nonlinear probe, allowing the direction to adapt at each step. This enables \ours{} to follow a curved trajectory toward the desired activation region, instead of relying on a single global direction or a fixed two-dimensional rotation plane for all activations.

%% file: content/exp.tex
\section{Experiments}
\input{table/table_1}

In this section, we evaluate \ours{} across three representative activation-steering tasks: helpfulness, truthfulness, and detoxification. These tasks test whether the proposed method can steer model behavior toward desirable attributes while preserving generation quality.

\noindent
\textbf{Base LLMs.}
We conduct experiments on four open-source language models with comparable parameter scales: Falcon-7B~\citep{almazrouei2023falcon}, Mistral-7B-v0.3~\citep{jiang2023mistral7b}, LLaMA-3.1-8B~\citep{meta2024llama3.1}, and Qwen2.5-7B~\citep{qwen2.5}. Additional details on the model implementations are provided in Appendix~\ref{a:model_detail}.

\noindent
\textbf{Baselines.}
We compare \ours{} against a diverse set of activation-steering baselines. These include Representation Engineering (RepE)~\citep{zou2023representation}, Inference-Time Intervention (ITI)~\citep{li2023inference}, Contrastive Activation Addition (CAA)~\citep{rimsky2024steering}, Minimally Modified Counterfactuals (MiMiC)~\citep{singh2024representation}, Linear-AcT~\citep{rodriguez2025controlling}, and ODESteer~\citep{zhao2026odesteer}. We also include norm-preserving methods that are most closely related to our setting, including HPR~\citep{pham2024hpr}, Spherical Steering~\citep{you2026spherical}, and Angular Steering~\citep{vu2026angular}. To ensure a controlled comparison, all methods are evaluated under the same steering protocol: we use the same steering layer and apply the intervention to newly generated tokens, following common practice in prior activation-steering evaluations~\citep{wehner2025taxonomy,bartoszcze2025representation}. Additional details on the selected steering layers are provided in Appendix~\ref{a:steering_position}.

\noindent
\textbf{Scope of Comparison.}
We focus on methods that are directly compatible with single-layer inference-time activation steering. Therefore, we do not include methods designed for substantially different goals, such as multi-attribute control~\citep{nguyen2025multi}, differential privacy~\citep{goel2025differentially}, or instruction-following improvement~\citep{stolfo2025improving}. We also exclude SADI~\citep{wang2025semanticsadaptive}, as it requires interventions across all layers and therefore does not fit our single-layer evaluation setup.

\noindent
\textbf{Datasets and Metrics.}
We evaluate on benchmarks corresponding to the three steering objectives. For helpfulness, we use UltraFeedback~\citep{cui2023ultrafeedback} and report the win rate against responses from the original model as the main metric, along with mean reward and 90th-percentile reward. For truthfulness, we evaluate on TruthfulQA~\citep{lin2021truthfulqa}, using Truth $\times$ Info as the primary metric and reporting Truth and Info separately for completeness. For detoxification, we use RealToxicityPrompts~\citep{gehman2020realtoxicityprompts} and report toxicity as the main metric. We additionally report perplexity and Dist-$n$ to measure generation quality and diversity. Further details on the datasets and evaluation metrics are provided in Appendix~\ref{a:dataset}.

\subsection{Experimental Results}

Table~\ref{tab:main} reports the main results across helpfulness, truthfulness, and detoxification. Overall, \ours{} consistently achieves the best performance on the primary metrics across all evaluated models. Specifically, \ours{} obtains the highest win rate on UltraFeedback, the highest truthfulness $\times$ informativeness score on TruthfulQA, and the lowest toxicity score on RealToxicityPrompts. These results show that \ours{} can effectively steer model behavior toward desirable attributes while maintaining strong generation quality.
The improvement over additive and linear steering methods, such as CAA, MiMiC, ITI, and Linear-AcT, suggests the importance of respecting the geometry of the activation space. Since these methods typically apply a fixed Euclidean intervention, they may distort the activation representation or move it away from the model's learned activation manifold. In contrast, \ours{} performs steering directly on the hypersphere, preserving the activation norm throughout the update.
Compared with existing norm-preserving methods, including HPR, Angular Steering, and Spherical Steering, \ours{} further benefits from its adaptive geodesic trajectory. Rather than assuming a fixed transformation or a single rotation path, \ours{} recomputes the steering direction at each intermediate point. By dividing a large steering step into multiple smaller geodesic updates, \ours{} can follow a more flexible path toward the desired activation region, leading to stronger and more consistent performance across models and tasks.

\subsection{Ablation Studies}

\input{table/table_2}

We conduct ablation studies to analyze the effect of three important design choices in \ours{}: the number of geodesic steps, the total steering strength, and the objective function design. All ablations are evaluated on TruthfulQA using True $\times$ Info as the primary metric.

\noindent
\textbf{Effect of the Number of Steps.}
Figure~\ref{fig:ablation_steps} studies how the number of geodesic steps affects performance when the total steering strength is fixed. Increasing the number of steps, or equivalently reducing the step size, brings a modest initial gain before the performance stabilizes. This suggests that a moderate number of steps is sufficient for reliable integration, and that \ours{} is not sensitive to the exact step-size choice. The stable trend also shows that the objective function provides a reliable steering direction along the path. Rather than relying on one large intervention, \ours{} repeatedly recomputes the direction at intermediate points, allowing the activation to move more smoothly on the sphere.

\noindent
\textbf{Effect of Steering Strength.}
Figure~\ref{fig:ablation_T} shows the effect of the total steering strength $T$, with the number of steps fixed to 15. Overall, \ours{} is not overly sensitive to small changes in $T$, and the performance remains strong within a reasonable range. When $T$ is too small, the intervention is too weak to sufficiently steer the model, leading to limited gains. In contrast, when $T$ is too large, the intervention can overly perturb the activation and degrade generation quality, which in turn reduces performance.
Different models still exhibit slightly different preferred ranges. For Mistral-7B and LLaMA3.1-8B, performance improves as $T$ increases from 0.5 and remains strong around $T=0.64$. Falcon-7B is more stable across the tested range, with smaller performance variation. Based on these results, we set $T=0.64$ in the main experiments, as it provides a strong and stable setting across models without requiring model-specific tuning.
\input{table/table_3}

\noindent
\textbf{Effect of Objective Function Design.}
Table~\ref{tab:ablation_objective} studies how the choice of objective function affects the steering performance. We compare a simple linear objective with two nonlinear alternatives: Random Fourier Features (RFF) \cite{rahimi2007random} and Polynomial Count Sketch. The linear objective already achieves strong information preservation, with an Info score of 96.2, but its True score is relatively lower. This suggests that a purely linear objective can preserve the original generation quality, but is less effective at capturing the nonlinear separation between desired and undesired activation regions.
Introducing nonlinear objectives improves the True score while maintaining a similar Info score. RFF increases the True score from 66.2 to 71.4, leading to a clear improvement in the combined True $\times$ Info metric. Polynomial Count Sketch performs the best among the three objectives, achieving 73.3 on True and 69.4 on True $\times$ Info, while keeping Info nearly unchanged. These results indicate that modeling nonlinear relationships in the activation space provides a more effective steering direction without sacrificing informativeness.

% \subsection{Runtime analysis}
% We further evaluate the computational overhead of \ours{} during autoregressive generation. The experiment measures generation throughput in tokens per second. As shown in Table~\ref{tab:runtime_analysis}, the baseline model achieves 52.4 tokens/s, while CAA and Spherical Steering introduce only small overheads of 1.2\% and 2.8\%, respectively.
% GeoSteer achieves 49.9 tokens/s, corresponding to a 4.7\% overhead compared with the baseline. This slight decrease is expected, since \ours{} performs iterative geodesic updates rather than applying a single linear intervention. However, the additional cost remains modest in practice. The results show that GeoSteer improves steering effectiveness while preserving most of the original generation throughput, making it practical for autoregressive inference.

\subsection{Runtime analysis}
We further evaluate the computational overhead of \ours{} during autoregressive generation. The experiment measures generation throughput in tokens per second on 4 NVIDIA A100 40GB GPUs across three backbone models. As shown in Table~\ref{tab:runtime_analysis}, the original models achieve 104.2, 96.1, and 104.6 tokens/s on Falcon-7B, Mistral-7B, and LLaMA3.1-8B, respectively.
\ours{} achieves 103.9, 95.5, and 102.8 tokens/s on the same backbones, corresponding to relative overheads of 0.3\%, 0.6\%, and 1.7\%, respectively. On average, \ours{} retains 100.7 tokens/s compared with 101.6 tokens/s for the original models, resulting in an average overhead of only 0.9\%. This small decrease is expected because \ours{} performs iterative geodesic updates during generation. Nevertheless, the overhead remains negligible in practice and is comparable to other efficient steering methods. These results show that \ours{} preserves almost all of the original generation throughput, making it practical for autoregressive inference.

%% file: table/table_1.tex
\begin{table*}[t!]
\centering
% \caption{Comparison of methods on Falcon-7B, Mistral-7B, LLaMA3.1-8B for helpfulness, truthfulness, and detoxification. 
% For helpfulness: ``Win'' is win rate, ``RM$_{\text{mean}}$'' is mean reward, and ``RM$_{\text{P90}}$'' is 90th percentile reward. 
% For truthfulness: ``T$\times$I'' is Truthfulness $\times$ Informativeness, with ``True'' and ``Info'' reported separately. 
% For detoxification: ``PPL'' is perplexity. 
% Results are averaged over three runs. 
% \textbf{Primary metrics are highlighted in \hl{blue}}; best and second-best are in \textbf{bold} and \underline{underline}. 
% % Dist-1/3 scores for detoxification are provided in Appendix~\ref{subapp:dists}.
% }
\caption{Comparison across Falcon-7B, Mistral-7B, and LLaMA3.1-8B on helpfulness, truthfulness, and detoxification.
For helpfulness, ``Win'' measures the win rate over the original model, and ``RM$_{\text{mean}}$'' / ``RM$_{\text{P90}}$'' report the mean and 90th-percentile reward.
For truthfulness, ``T$\times$I'' denotes Truthfulness $\times$ Informativeness, with ``True'' and ``Info'' shown separately.
For detoxification, ``PPL'' denotes perplexity.
Results are averaged over five runs.
Primary metrics are highlighted in \hl{blue}
% For the reported main metrics and selected auxiliary metrics, the best and second-best results are marked by \textbf{bold} and \underline{underline}.
; for main and selected auxiliary metrics, the best and second-best results are marked by \textbf{bold} and \underline{underline}.
}
\large
\renewcommand{\arraystretch}{1.2}
\resizebox{\linewidth}{!}{%

\begin{tabular}{cc|
>{\columncolor{lightpurple!25}}c c c|
>{\columncolor{lightpurple!25}}c c c|
>{\columncolor{lightpurple!25}}c c c}

\toprule[1.5pt]
\multirow{2}{*}{\textbf{Method}} & \multirow{2}{*}{\textbf{Model}} &
\multicolumn{3}{c|}{\textbf{Helpfulness} \textsubscript{\scriptsize\textbf{(Ultrafeedback)}}} &
\multicolumn{3}{c|}{\textbf{Truthfulness} \textsubscript{\scriptsize\textbf{(TruthfulQA)}}} &
\multicolumn{3}{c}{\textbf{Detoxification} \textsubscript{\scriptsize\textbf{(Real Toxicity Prompts)}}} \\
\cmidrule(lr){3-5}\cmidrule(lr){6-8}\cmidrule(lr){9-11}
& & \textbf{Win (\%)}\,$\boldsymbol{\uparrow}$ & \textbf{RM$_{\text{mean}}$}\,$\boldsymbol{\uparrow}$ & \textbf{RM$_{\text{P90}}$}\,$\boldsymbol{\uparrow}$ &
\textbf{T$\times$I (\%)}\,$\boldsymbol{\uparrow}$ & \textbf{True (\%)}\,$\boldsymbol{\uparrow}$ & \textbf{Info (\%)}\,$\boldsymbol{\uparrow}$ &
\textbf{ Toxic }\,$\boldsymbol{\downarrow}$ & \textbf{PPL}\,$\boldsymbol{\downarrow}$ & \textbf{Dist-2}\,$\boldsymbol{\uparrow}$ \\
\midrule

% ---------------- Falcon-7B ----------------
Original     & \multirow{11}{*}{\rotatebox{90}{Falcon-7B }} & 50.0 & -12.5 & -2.9 & 28.4 & 31.3 & 96.3 & 0.181 & 15.8 & 94.4 \\
\midrule
RepE         &  & 49.7 & -12.9 & -2.8 & 28.7 & 31.6 & 96.2 & 0.187 & 16.3 & 94.7 \\
ITI          &  & 48.5 & -12.5 & -2.9 & 29.4 & 32.5 & 96.1 & 0.187 & 15.6 & 94.5 \\
CAA          &  & \underline{54.0} & -12.4 & \textbf{-1.2} & 34.1 & 37.6 & 96.0 & 0.093 & 17.1 & 94.7 \\
MiMiC        &  & 45.5 & -13.1 & -3.4 & 35.4 & 43.9 & 90.1 & 0.167 & 16.5 & 94.1 \\
Linear-AcT   &  & 49.8 & -12.8 & -2.5 & 34.1 & 37.7 & 96.1 & 0.168 & 15.9 & 94.6 \\
ODESteer     &  & 49.0 & -12.8 & -3.4 & 33.3 & 36.2 & 96.6 & \underline{0.088} & 37.5 & 93.4 \\
HPR          &  & 48.9 & -13.5 & -3.5 & 34.9 & 39.2 & 95.7 & 0.099 & 43.1 & 92.9 \\
Angular Steering   &  & 51.5 & -12.5 & -2.9 & \underline{39.4} & 43.7 & 94.7 & 0.113 & 16.6 & 96.0 \\
Spherical Steering    &  & 51.8 & \underline{-12.1} & -2.3 & 33.4 & \underline{45.2} & 86.7 & 0.101 & 19.0 & 93.8 \\
\midrule
\textbf{\ours{} (Ours)} & & \textbf{57.3} & \textbf{-11.8} & \underline{-2.0} & \textbf{41.1} & \textbf{46.5} & 93.6 & \textbf{0.071} & 17.9 & 93.5 \\
\midrule[1pt]
\midrule[1pt]

% ---------------- Mistral-7B ----------------
Original     & \multirow{11}{*}{\rotatebox{90}{Mistral-7B }} & 50.0 & -11.3 & -1.2 & 36.7 & 40.8 & 95.6 & 0.125 & 18.7 & 98.9 \\
\midrule
RepE         &  & 36.6 & -14.5 & -3.4 & 33.3 & 43.8 & 88.9 & 0.116 & 39.9 & 93.2 \\
ITI          &  & 47.0 & -11.2 & -1.0 & 42.6 & 48.3 & 94.0 & 0.060 & 22.2 & 99.0 \\
CAA          &  & \underline{53.4} & \underline{-9.6} & 0.6 & 53.2 & 58.6 & 94.5 & 0.041 & 20.8 & 99.0 \\
MiMiC        &  & 49.0 & -11.4 & -0.8 & 40.0 & 47.1 & 92.7 & 0.103 & 18.9 & 98.9 \\
Linear-AcT   &  & 51.1 & -10.9 & -0.1 & 42.6 & 47.6 & 95.0 & 0.102 & 19.3 & 98.9 \\
ODESteer     &  & 53.3 & -10.5 & \underline{0.7} & 51.5 & 56.9 & 94.5 & \underline{0.034} & 20.2 & 99.3 \\
HPR          &  & 52.3 & -10.1 & 0.7 & 50.4 & 56.4 & 89.4 & 0.058 & 43.1 & 92.9 \\
Angular Steering   &  & 50.4 & -11.1 & -0.7 & 51.0 & 58.0 & 93.0 & 0.049 & 31.1 & 99.2 \\
Spherical Steering    &  & 47.3 & -11.9 & -2.4 & \underline{58.1} & \underline{69.9} & 86.8 & 0.044 & 34.3 & 98.2 \\
\midrule
\textbf{\ours{} (Ours)} & & \textbf{54.7} & \textbf{-8.9} & \textbf{0.9} & \textbf{63.2} & \textbf{70.0} & 93.0 & \textbf{0.015} & 39.6 & 99.1 \\
\midrule[1pt]
\midrule[1pt]

% ---------------- LLaMA3.1-8B ----------------
Original     & \multirow{11}{*}{\rotatebox{90}{LLaMA3.1-8B }} & 50.0 & -10.5 & -0.3 & 41.5 & 44.1 & 96.7 & 0.124 & 18.7 & 99.0 \\
\midrule
RepE         &  & 49.4 & -10.4 & -1.2 & 40.6 & 43.3 & 96.7 & 0.143 & 38.5 & 96.4 \\
ITI          &  & 48.8 & -10.5 & -0.5 & 43.7 & 46.1 & 96.9 & 0.087 & 19.7 & 98.6 \\
CAA          &  & 57.6 & -8.9 & 1.5 & 50.1 & 53.0 & 97.6 & 0.042 & 21.5 & 99.2 \\
MiMiC        &  & 57.8 & -9.2 & 1.1 & 54.1 & 60.8 & 92.5 & 0.098 & 19.2 & 99.3 \\
Linear-AcT   &  & 52.8 & -9.9 & 0.9 & 51.5 & 53.9 & 97.1 & 0.103 & 18.7 & 98.9 \\
ODESteer     &  & \underline{59.0} & -8.5 & 1.6 & \underline{67.7} & 72.6 & 94.7 & 0.032 & 24.6 & 98.7 \\
HPR          &  & 52.2 & -10.0 & 0.7 & 57.0 & 60.7 & 94.0 & 0.040 & 36.1 & 97.5 \\
Angular Steering  &  & 55.7 & \underline{-8.2} & \underline{1.7} & 67.1 & 71.2 & 95.5 & 0.043 & 20.6 & 99.1 \\
Spherical Steering    &  & 52.3 & -9.5 & 0.6 & 59.5 & \textbf{88.4} & 70.1 & \underline{0.028} & 371.5 & 95.2 \\
\midrule
\textbf{\ours{} (Ours)} & & \textbf{63.2} & \textbf{-7.4} & \textbf{2.1} & \textbf{69.4} & \underline{73.3} & 95.8 & \textbf{0.018} & 25.7 & 99.4 \\
\midrule[1pt]
\midrule[1pt]

% ---------------- Qwen2.5-7B ----------------
Original     & \multirow{11}{*}{\rotatebox{90}{\diff{Qwen2.5-7B}}} & 50.0 & -5.4 & 5.2 & 63.8 & 78.2 & 85.2 & 0.097 & 21.2 & 99.2 \\
\midrule
RepE         &  & 50.7 & -5.4 & 5.6 & 62.5 & 77.3 & 85.1 & 0.109 & 21.9 & 99.1 \\
ITI          &  & 48.8 & -5.6 & \underline{5.8} & 65.4 & 78.9 & 86.3 & 0.097 & 21.2 & 99.1 \\
CAA          &  & 49.8 & -5.4 & 5.1 & 66.3 & 80.0 & 85.8 & 0.049 & 20.3 & 99.3 \\
MiMiC        &  & 48.1 & -6.2 & 4.0 & 62.8 & 82.1 & 79.9 & 0.091 & 21.2 & 99.1 \\
Linear-AcT   &  & 47.5 & -5.9 & 5.3 & 63.6 & 79.3 & 83.9 & 0.084 & 22.1 & 99.2 \\
ODESteer    &  & \underline{51.4} & \underline{-5.4} & 5.7 & 63.7 & 79.3 & 84.1 & 0.087 & 21.8 & 99.0 \\
HPR          &  & 48.4 & -5.9 & 3.8 & 65.6 & 77.9 & 84.3 & 0.046 & 23.5 & 99.1 \\
Angular Steering   &  & 46.6 & -6.2 & 4.7 & \underline{66.3} & \textbf{85.6} & 80.3 & 0.049 & 21.4 & 99.2 \\
Spherical Steering    & & 17.2 & -15.8 & -3.0 & 61.2 & 81.7 & 78.6 & \underline{0.034} & 125.9 & 97.8 \\
\midrule
\textbf{\ours{} (Ours)} & & \textbf{53.2} & \textbf{-4.6} & \textbf{6.2} & \textbf{68.1} & \underline{82.3} & 85.2 & \textbf{0.029} & 22.1 & 99.2 \\
\bottomrule[1.5pt]
\end{tabular}
}
\label{tab:main}
\end{table*}

%% file: table/table_2.tex
\begin{figure*}[t!]
    \centering

    \begin{subfigure}[t]{0.48\linewidth}
        \centering
        \includegraphics[width=\linewidth]{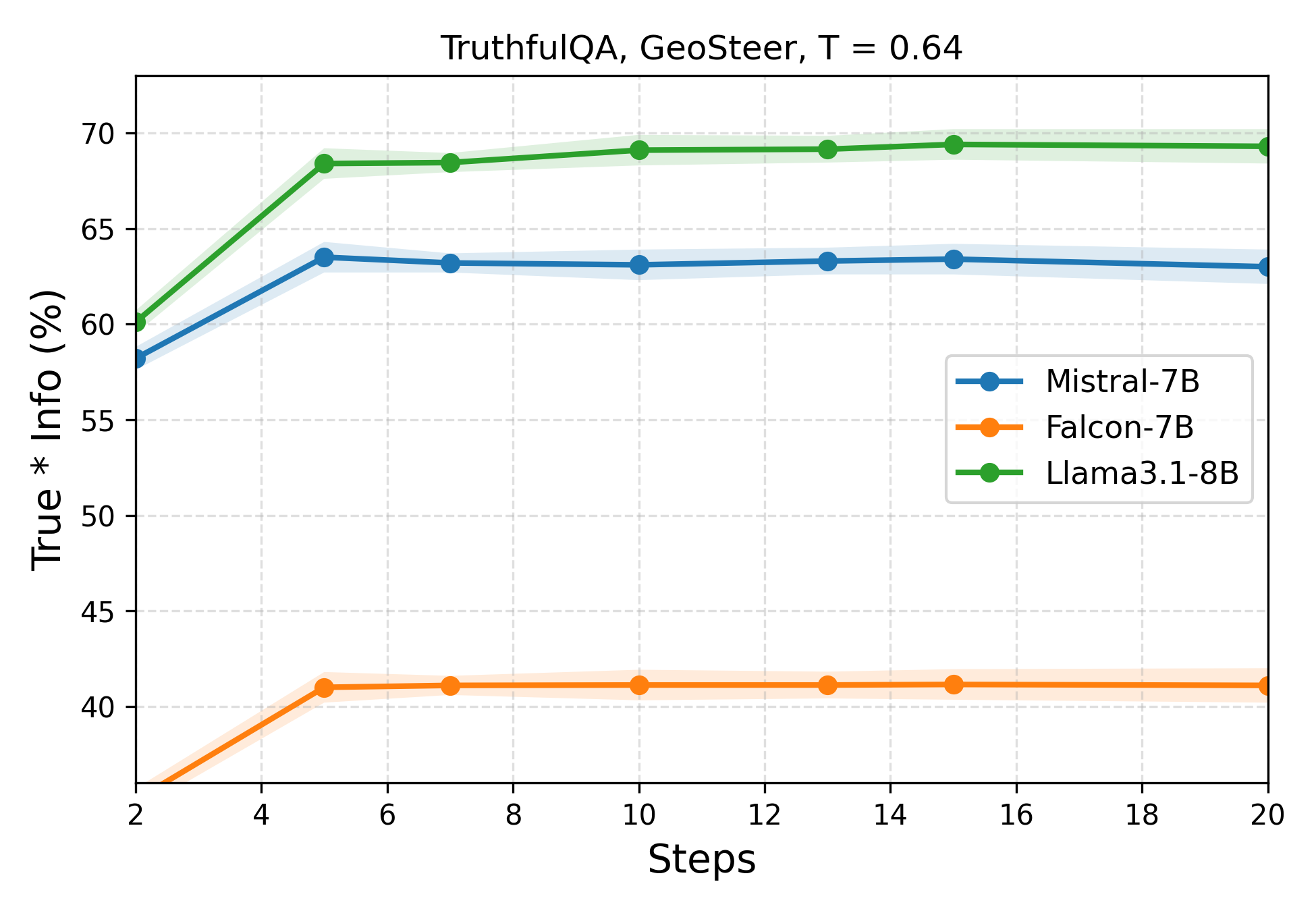}
        \caption{Effect of the number of geodesic steps.}
        \label{fig:ablation_steps}
    \end{subfigure}
    \hfill
    \begin{subfigure}[t]{0.48\linewidth}
        \centering
        \includegraphics[width=\linewidth]{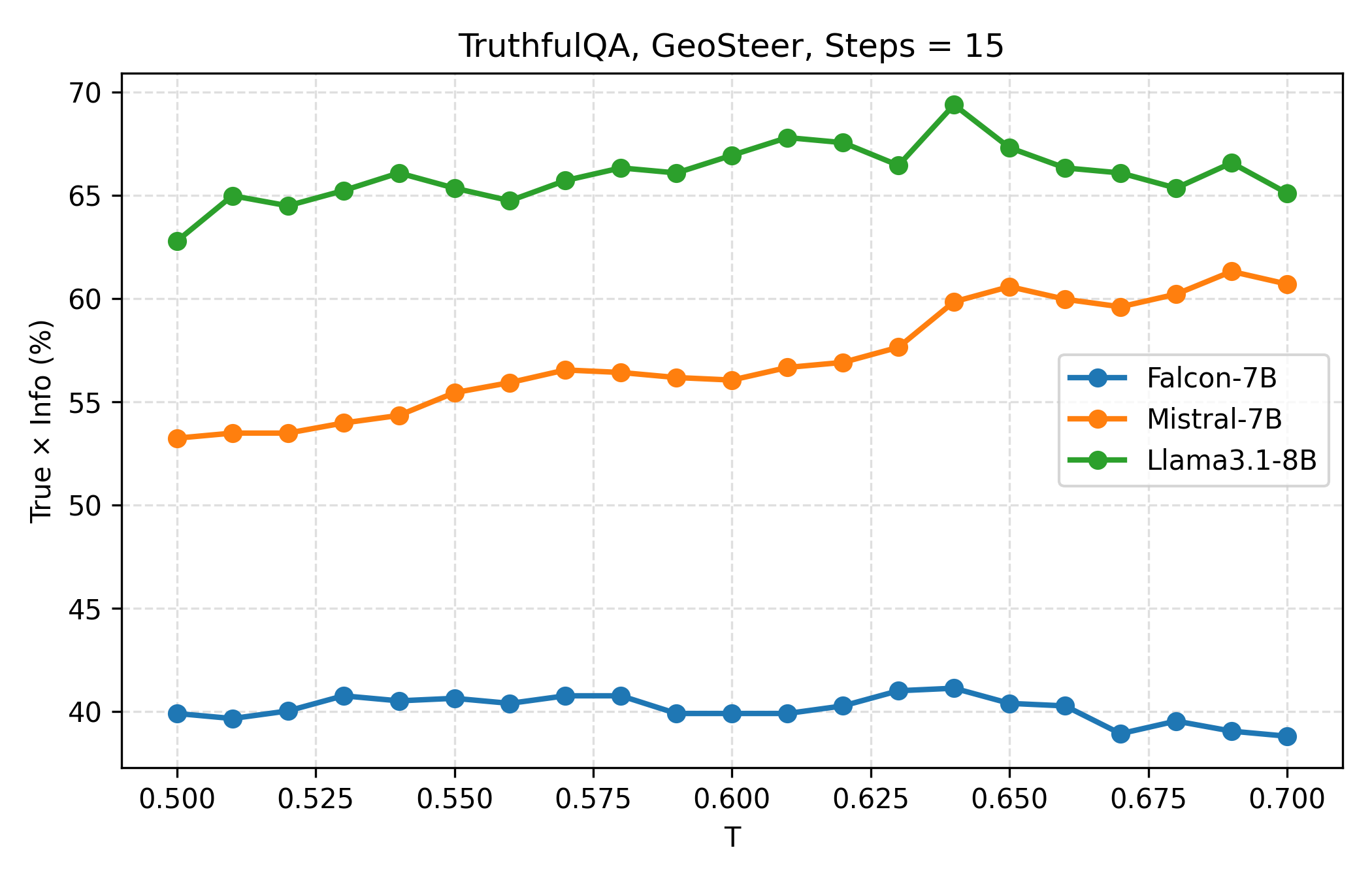}
        \caption{Effect of total steering strength $T$.}
        \label{fig:ablation_T}
    \end{subfigure}

    \caption{
    Ablation studies on TruthfulQA using True $\times$ Info as the main metric.
    We analyze the effect of the number of geodesic steps and the total steering strength $T$.
    }
    \label{fig:ablation}
\end{figure*}

%% file: table/table_3.tex
% \begin{table}[t]
% \centering
% \caption{Runtime analysis on LLaMA3.1-8B. Tokens/s measures autoregressive generation speed, and overhead is computed relative to the unsteered baseline.}
% \resizebox{0.96\columnwidth}{!}{%
% \label{tab:runtime_analysis}
% \begin{tabular}{lcc}
% \toprule
% \textbf{Method} & \textbf{Tokens/s} $\uparrow$ & \textbf{Overhead} $\downarrow$ \\
% \midrule
% Baseline & 52.4 & - \\
% CAA      & 51.8 & 1.2\% \\
% Spherical Steering   & 50.9 & 2.8\% \\
% GeoSteer & 49.9 & 4.7\% \\
% \bottomrule
% \end{tabular}
% }
% \end{table}

\begin{table}[t]
\centering
\caption{Runtime analysis across different backbone models. Tokens/s measures autoregressive generation speed.}
\label{tab:runtime_analysis}
\resizebox{\columnwidth}{!}{%
\begin{tabular}{lccc}
\toprule
\textbf{Method} & \textbf{Falcon-7B} & \textbf{Mistral-7B} & \textbf{LLaMA3.1-8B} \\
\midrule
Original & 104.2 & 96.1 & 104.6 \\
\midrule
RepE & 104.0 & 95.7 & 102.9 \\
ITI & 104.0 & 95.7 & 103.0 \\
CAA & 104.1 & 95.8 & 103.2 \\
MiMiC & 101.0 & 93.9 & 100.8 \\
Linear-AcT & 104.1 & 95.9 & 103.3 \\
ODESteer & 103.8 & 95.5 & 102.9 \\
Angular Steering & 104.1 & 95.7 & 103.0 \\
Spherical Steering & 104.1 & 95.6 & 103.1 \\
\textbf{\ours{} (Ours)} & 103.9 & 95.5 & 102.8 \\
\bottomrule
\end{tabular}
}
\end{table}

\begin{table}[t]
\centering
\caption{
Ablation study on the objective function design using LLaMA3.1-8B.
We compare a linear objective against two nonlinear alternatives based on Random Fourier Features (RFF) and Polynomial Count Sketch.
}
\resizebox{\columnwidth}{!}{%
\label{tab:ablation_objective}
\begin{tabular}{lccc}
\toprule
\textbf{Objective Function} & \textbf{True} $\uparrow$ & \textbf{Info} $\uparrow$ & \textbf{True $\times$ Info} $\uparrow$ \\
\midrule
Linear & 66.2 & 96.2 & 62.4 \\
RFF & 71.4 & 96.0 & 67.8 \\
Polynomial Count Sketch & 73.3 & 95.8 & 69.4 \\
\bottomrule
\end{tabular}
}
\end{table}

%% file: content/concl.tex
\section{Conclusion}

In this work, we introduced \ours{}, an optimization-based method for norm-preserving activation steering. Unlike existing methods that rely on predefined one-step edits, \ours{} formulates steering as a Riemannian optimization problem and moves activations through a sequence of small geodesic steps on the activation sphere. 
% At each step, the steering direction is adaptively recomputed using an objective function based on the log-density ratio between positive and negative activations.
At each step, the steering direction is adaptively recomputed from the local gradient of a learned nonlinear activation-space objective.
This design better respects the geometry of normalized representations and provides a smoother and more stable way to control model behavior at inference time. Across TruthfulQA, RealToxicityPrompts, and UltraFeedback, \ours{} consistently improves over strong activation steering baselines. Our ablation studies further show the importance of the number of geodesic steps, the total steering strength, and the nonlinear objective function design, while the runtime analysis demonstrates that the additional cost remains modest during autoregressive generation.

% \noindent
% \textbf{Limitations}. First, \ours{} currently focuses on steering directions obtained from contrastive activation differences and does not incorporate other classes of direction-discovery methods, such as sparse autoencoders (SAEs). Extending \ours{} to SAE-based features is non-trivial because the feature space can be very high-dimensional and sparse, making it unclear how to define stable geodesic directions and integrate them efficiently. Second, our current method applies steering at a single selected layer. While this provides a clean and efficient intervention setup, different layers may encode complementary behavioral or semantic information. As future work, we plan to investigate how to combine \ours{} with SAE-based direction discovery and how to extend geodesic steering to multiple layers in a principled and computationally efficient manner.

\section{Limitations}

Although \ours{} provides an effective framework for norm-preserving activation steering, it has several limitations.

\begin{itemize}
    \item \textbf{Integration with other direction-discovery methods.}
    Our current implementation learns the steering objective from contrastive activation examples. This makes the method simple and compatible with standard activation-steering pipelines, but it does not directly incorporate other direction-discovery mechanisms, such as sparse autoencoders (SAEs). Extending \ours{} to SAE-based features is non-trivial because SAE representations are often high-dimensional, sparse, and feature-wise rather than direction-wise. It is therefore unclear how to select stable feature combinations, convert them into smooth tangent directions on the activation sphere, and perform geodesic updates efficiently without introducing noise or instability.

    % \item \textbf{Single-layer steering.}
    % \ours{} currently applies steering at a single selected layer. This design keeps the intervention efficient and makes comparisons with prior single-layer steering methods straightforward. However, model behavior may be distributed across multiple layers, with different layers encoding complementary lexical, factual, and semantic information. Steering only one layer may therefore miss useful signals from other parts of the network. A natural extension is to develop a multi-layer version of GeoSteer in which layer selection, steering strength, and geodesic updates are coordinated jointly rather than chosen independently.

    \item \textbf{Potential risks and unintended behavior.}
    Activation steering is a general-purpose inference-time control mechanism. Although our experiments focus on desirable behaviors, such as improving truthfulness, reducing toxicity, and improving preference-oriented generation, the same type of method could be misused with inappropriate objectives or steering directions to induce undesirable behaviors, including toxic, biased, deceptive, or otherwise harmful outputs. In addition, preserving the activation norm does not guarantee that all unintended behavioral changes are avoided. Therefore, activation steering methods should be carefully evaluated for downstream safety, robustness, and misuse risks before deployment.
\end{itemize}

As future work, we plan to investigate stable integration of \ours{} with SAE-based feature discovery and to conduct broader safety evaluations that better characterize unintended behavioral changes and misuse risks across tasks and model families.

% As future work, we plan to investigate both directions: integrating \ours{} with SAE-based feature discovery and extending geodesic steering to multiple layers in a stable and computationally efficient manner.

%% file: content/appendix.tex
\clearpage
\section{Appendix}
\label{sec:appendix}

\subsection{LLM Usage}

In this work, Large Language Models (LLMs) were used only as writing and coding assistants. Specifically, they were used to help polish the manuscript for grammar, clarity, and readability, and to draft small portions of experimental code.

All LLM-assisted content was carefully reviewed, verified, and revised by the authors. The research ideas, theoretical framework, methodology, experimental design, and analysis were developed by the authors. LLMs were not used to generate the core scientific contributions or to draw conclusions from the experimental results.

The authors take full responsibility for the correctness of the theoretical claims, the validity of the experiments, and the reported results. Any text or code produced with LLM assistance was checked to ensure correctness, originality, and compliance with ethical research standards.

\subsection{Scientific Artifacts}
\label{app:artifacts}

Our experiments use only publicly available scientific artifacts, including benchmark datasets, pretrained language models, evaluation models, and baseline implementations. The datasets used in our evaluation include TruthfulQA, RealToxicityPrompts, and UltraFeedback/UltraFeedback Binarized. TruthfulQA is used for truthfulness evaluation, RealToxicityPrompts is used for toxicity-oriented generation evaluation, and UltraFeedback is used for preference-oriented evaluation. These datasets are English-language text benchmarks and are used only for research evaluation. We report aggregate results and do not release newly collected human-subject data.

We use all artifacts in accordance with their respective licenses and terms of use. TruthfulQA and RealToxicityPrompts are released under Apache-2.0, while UltraFeedback/UltraFeedback Binarized is released under the MIT license. The pretrained language models used in our experiments are also publicly available: Falcon-7B, Mistral-7B, and Qwen2.5-7B are released under Apache-2.0, while LLaMA3.1-8B is released under the LLaMA 3.1 Community License. We use these models only as research artifacts for controlled evaluation of activation-steering methods.

Our use of these artifacts is consistent with their intended research and evaluation purposes. We do not redistribute modified versions of the datasets or pretrained models as part of this work.

We use publicly available benchmark datasets and do not collect new human-subject data. Some datasets used in our experiments may contain sensitive or offensive text by construction, especially RealToxicityPrompts, which is designed for evaluating toxic language generation. We use these datasets only for research evaluation, do not redistribute modified versions of the data, and report aggregate evaluation results rather than individual examples. We do not release any newly collected personally identifying information.

\subsection{PCS Hyperparameters}
\label{a:pcs}

As described in \ref{sec:geosteer}, we use Polynomial Count Sketch~\citep{pham2013fast} to construct randomized polynomial features. This provides an efficient approximation to the polynomial kernel
\begin{equation}
K(x,y) = (\gamma x^\top y + c_0)^d,
\end{equation}
where $\gamma$ controls the scale of the inner product, $c_0$ is the constant offset, and $d$ denotes the polynomial degree. The approximation also depends on the number of random features, denoted by $N_{\text{poly}}$. We set $\gamma=0.1$, $c_0=1.0$, $d=2$, and $N_{\text{poly}}=8000$ in all experiments. We found this configuration to be stable across different datasets and base models.

\subsection{Norm Preservation of GeoSteer}
\label{a:norm}

We show that the GeoSteer update preserves the unit-norm constraint. At step $t$, assume that the current activation direction lies on the unit sphere:
\begin{equation}
\|z^{(t)}\|_2 = 1.
\end{equation}
The update direction $u^{(t)}$ is obtained by projecting the gradient onto the tangent space of the sphere and normalizing it. Therefore,
\begin{equation}
\|u^{(t)}\|_2 = 1,
\qquad
\left(z^{(t)}\right)^\top u^{(t)} = 0.
\end{equation}
GeoSteer updates the activation direction as
\begin{equation}
z^{(t+1)}
=
\cos(\eta)z^{(t)}
+
\sin(\eta)u^{(t)}.
\end{equation}
To verify that the updated point remains on the sphere, we compute its squared norm:
\begin{equation}
\begin{aligned}
\left\|z^{(t+1)}\right\|_2^2
&=
\left\|
\cos(\eta)z^{(t)}
+
\sin(\eta)u^{(t)}
\right\|_2^2 .
\end{aligned}
\end{equation}
Expanding the squared norm gives
\begin{equation}
\begin{aligned}
\left\|z^{(t+1)}\right\|_2^2
&=
\cos^2(\eta)
\left\|z^{(t)}\right\|_2^2
+
\sin^2(\eta)
\left\|u^{(t)}\right\|_2^2 \\
&\quad
+
2\cos(\eta)\sin(\eta)
\left(z^{(t)}\right)^\top u^{(t)} .
\end{aligned}
\end{equation}
Since $z^{(t)}$ and $u^{(t)}$ are unit vectors and are orthogonal, we obtain
\begin{equation}
\begin{aligned}
\left\|z^{(t+1)}\right\|_2^2
&=
\cos^2(\eta)
+
\sin^2(\eta) \\
&= 1.
\end{aligned}
\end{equation}
Thus,
\begin{equation}
z^{(t+1)} \in \mathbb{S}^{d-1}.
\end{equation}
Since the initial point $z^{(0)}=h/\|h\|_2$ is on the unit sphere, the same argument applies at every step. Therefore, all intermediate points remain on the sphere, and the final reconstructed activation
\begin{equation}
\hat{h}=\|h\|_2 z^{(K)}
\end{equation}
preserves the original activation norm.

\subsection{Steering Model Details}
\label{a:model_detail}

We conduct experiments using the following open-source language models:
\begin{itemize}
    \item Falcon-7B: \texttt{tiiuae/falcon-7b};
    \item Mistral-7B: \texttt{mistralai/Mistral-7B-v0.3};
    \item LLaMA3.1-8B: \texttt{meta-llama/Llama-3.1-8B};
    \item Qwen2.5-7B: \texttt{Qwen/Qwen2.5-7B}.
\end{itemize}

For all models and tasks, we use the same generation configuration: temperature is set to $0.7$, top-$p$ to $0.9$, and repetition penalty to $1.1$.

\subsection{Steering Position}
\label{a:steering_position}

\begin{figure}[t]
    \centering
    \includegraphics[width=\columnwidth]{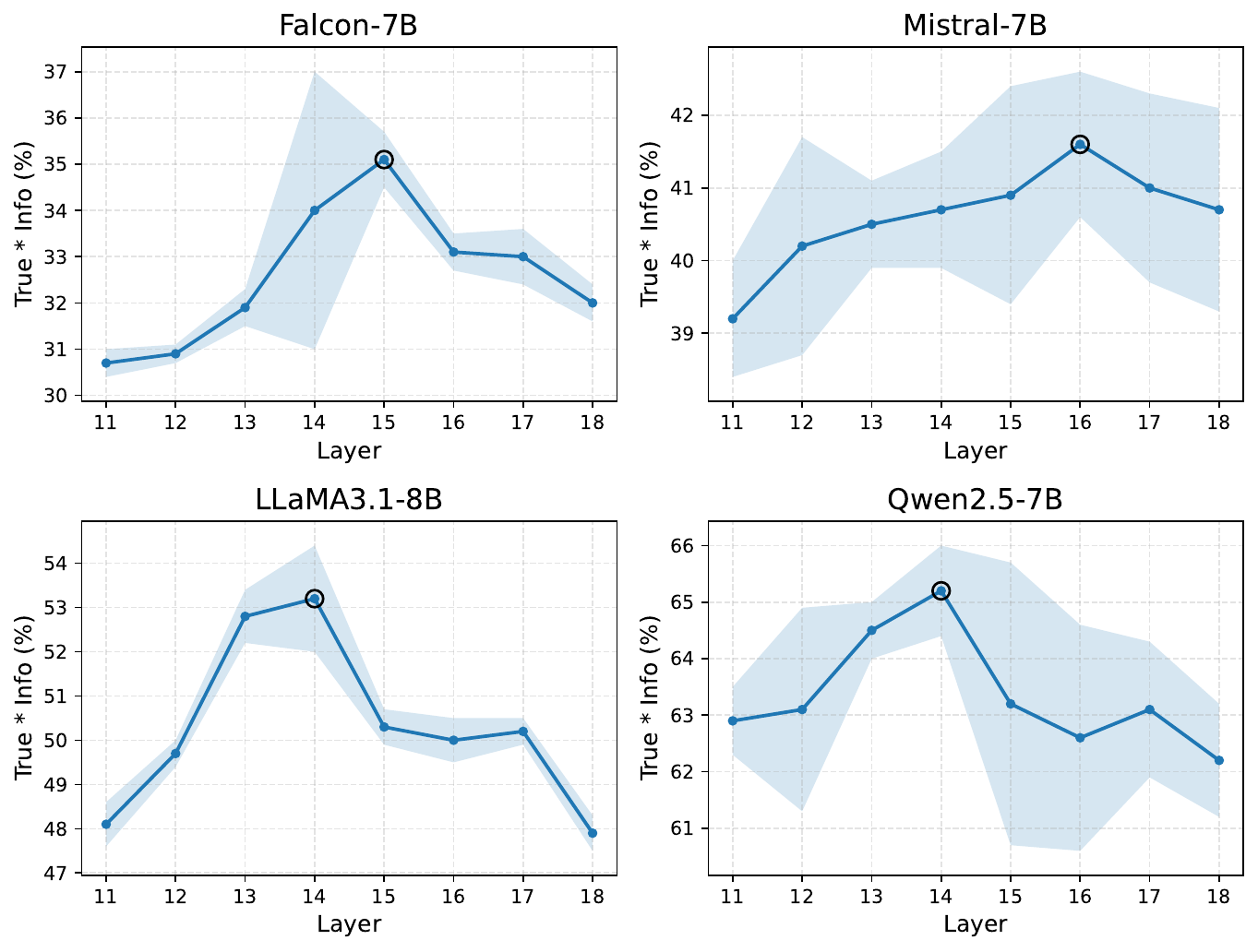}
    \caption{
    Layer-wise steering performance on TruthfulQA using CAA for layer selection.
    We report True$\times$Info across layers 11--18 for each base model.
    }
    \label{fig:layer_selection}
\end{figure}

For a fair comparison, we apply \ours{} and all baselines at the same residual-stream position within each model. Steering is applied to all newly generated tokens during decoding. To choose the steering layer, we first run CAA~\citep{rimsky2024steering} on TruthfulQA across all layers of each model and evaluate performance using the True$\times$Info metric. The layer-wise results are shown in Figure~\ref{fig:layer_selection}.

Based on this analysis, we use layer 15 for Falcon-7B, layer 16 for Mistral-7B, layer 14 for LLaMA3.1-8B, and layer 14 for Qwen2.5-7B. We use CAA only as a layer-selection tool, so that all methods are evaluated under the same steering position rather than tuning the layer separately for each method.

\subsection{Datasets}
\label{a:dataset}

\noindent
\textbf{UltraFeedback.}
We use the UltraFeedback Binarized dataset, where each prompt is associated with a preferred response and a rejected response. From this dataset, we construct 10k training pairs, 500 validation pairs, and 500 test prompts under three random seeds. For evaluation, we score model outputs using Skywork-Reward-V2-LLaMA-3.1-8B. We report the average reward score, denoted as RMmean, the 90th percentile reward score, denoted as RMP90, and the reward-model win rate relative to the baseline model.

\noindent
\textbf{Reward-Model Win Rate.}
Given a set of prompts $\{x_i\}_{i=1}^{N}$ and two systems $A$ and $B$, let $s_i^A$ and $s_i^B$ denote their reward-model scores on the same prompt. Following \cite{lambert-etal-2025-rewardbench}, we define the win rate of system $A$ over system $B$ as:

\begin{equation}
\begin{aligned}
\mathrm{Win}(A,B)
= \frac{1}{N}\sum_{i=1}^{N}
\Big[
&\mathbb{I}(s_i^A > s_i^B) \\
&+ \frac{1}{2}\mathbb{I}(s_i^A = s_i^B)
\Big].
\end{aligned}
\end{equation}

A win rate of $0.5$ indicates parity with the baseline, while values above $0.5$ indicate that system $A$ is preferred by the reward model more often than system $B$. Ties are counted as half wins.

\noindent
\textbf{TruthfulQA.}
For TruthfulQA, we follow the generation-based evaluation setting used by Li et al. (2023). The benchmark contains 817 questions, which are expanded into 5,918 question--answer pairs. We use 40\% of the data for training and 10\% for validation to select hyperparameters. We then conduct two-fold cross validation so that all TruthfulQA questions are included in the test split. The original TruthfulQA benchmark used two fine-tuned GPT-3 models as truthfulness and informativeness judges. Since these models are no longer publicly available, we instead use \texttt{allenai/truthfulqa-truth-judge-llama2-7B}, \texttt{allenai/truthfulqa-info-judge-llama2-7B} to evaluate truthfulness and informativeness, respectively.

\noindent
\textbf{RealToxicityPrompts.}
For detoxification, we use the Jigsaw Unintended Bias in Toxicity Classification dataset for training and RealToxicityPrompts \cite{gehman2020realtoxicityprompts} for testing. Specifically, we uniformly sample 10k sentences from the Jigsaw dataset according to toxicity scores, resulting in 5k toxic and 5k benign training examples. For evaluation, we select 500 toxic prompts from RealToxicityPrompts as inputs to the language models. We use \texttt{unitary/toxic-bert} to measure the toxicity of generated continuations. In addition, we report perplexity and Dist-$n$ scores to assess generation quality and diversity.

\noindent
\textbf{Activation Collection.}
For TruthfulQA and UltraFeedback, each training example contains a prompt paired with a positive and a negative response. We construct the model input by concatenating the prompt with each corresponding response and then forward the full sequence through the LLM. For the detoxification task, the Jigsaw dataset does not provide question--answer pairs; therefore, we directly use the toxic and non-toxic prompts as inputs for activation extraction. Following common practice in activation steering~\citep{wehner2025taxonomy}, we extract hidden states at the final token position of each input sequence to obtain positive and negative activations. This design aligns with the autoregressive decoding process, where steering is applied when predicting newly generated tokens.